\pdfoutput=1
\documentclass[11pt]{article}
\usepackage[]{ACL2023}
\usepackage{times}
\usepackage{latexsym}
\usepackage[T1]{fontenc}
\usepackage[utf8]{inputenc}
\usepackage{hyperref}
\usepackage{tcolorbox}
\usepackage{microtype}
\usepackage{inconsolata}

\title{A City of Millions: Mapping Literary Social Networks At Scale}

\author{Sil Hamilton$^{1*}$, Rebecca M. M. Hicke$^{2*}$, David Mimno$^{1}$, Matthew Wilkens$^{1}$ \\
  $^{1}$Department of Information Science \\
  $^{2}$Department of Computer Science \\
  Cornell University \\
  \texttt{\{srh255,rmh327,mimno,wilkens\}@cornell.edu} \\
  $^{*}$Equal contribution}

\begin{document}
\maketitle
\begin{abstract}
We release 70,509 high-quality social networks extracted from multilingual fiction and nonfiction narratives.
We additionally provide metadata for $\sim$30,000 of these texts (73\% nonfiction and 27\% fiction) written between 1800 and 1999 in 58 languages.
This dataset provides information on historical social worlds at an unprecedented scale, including data for 2,510,021 individuals in 2,805,482 pair-wise relationships annotated for affinity and relationship type.
We achieve this scale by automating previously manual methods of extracting social networks; specifically, we adapt an existing annotation task as a language model prompt, ensuring consistency at scale with the use of structured output.
This dataset serves as a unique resource for humanities and social science research by providing data on cognitive models of social realities.
\end{abstract}

\section{Introduction}
Literary scholars have long been interested in the social worlds of novels. 
Novels depict social configurations across time and space at varying levels of abstraction, from the grand descriptions of geopolitical intrigue in \emph{War and Peace} to the personal relationships underpinning \emph{In Search of Lost Time}.
While social networks cannot represent the full detail and nuance of literary works, they provide a uniform format to identify large-scale patterns.
However, prior attempts at extracting social networks from literary texts have been hindered by a dependency on supervised machine learning models limited in accuracy and scalability.

\begin{figure}[t]
    \centering
    \includegraphics[width=0.95\linewidth]{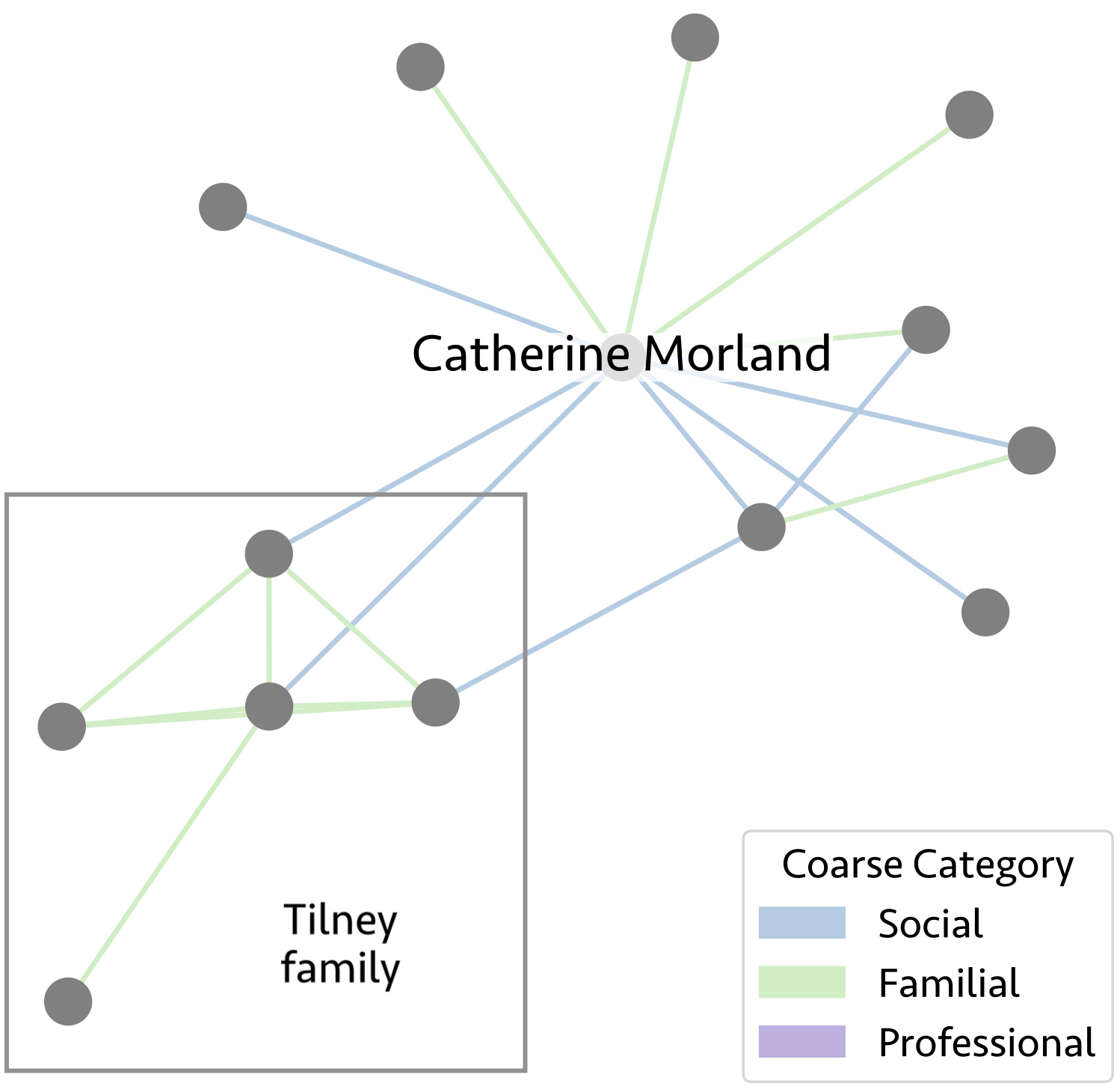}
    \caption{The graph of relationships in \textit{Northanger Abbey} by Jane Austen created by our model. Note the presence of the Tilney family at the bottom left of the figure.}
    \label{fig:northanger_abbey}
\end{figure}

In this work, we present a dataset of high quality social networks extracted from 70,509 literary texts, such as that shown in Figure \ref{fig:northanger_abbey}. 
We extract the networks, including affinities and relationship types, using a novel method that passes a modified prompt from \citet{massey2015annotating} to \texttt{Gemini 1.5 Flash} configured to output JSON. 
We validate this approach by demonstrating that it produces annotations similar to the manual annotations provided by \citet{massey2015annotating}.
In addition to networks, we also provide extended metadata for a subset of $\sim$30,000 works, of which 22,015 are nonfiction and 7,331 fiction.

This dataset will provide researchers with the opportunity to evaluate literary and social hypotheses at scale.
As an initial example, we show that nonfiction networks consist of more communities and are less clustered than fiction networks. 
This may help explain why characters in non-fiction texts travel more \cite{wilkens2024small} as, intuitively, a text consisting of fewer social communities may feature fewer locations.\footnote{We provide the full dataset \href{https://github.com/srhm-ca/pgn}{here.}}

\section{Related Work}

\paragraph{Literary social network extraction.}

Significant previous research has addressed extracting social networks from literary texts. One traditional approach involves creating networks by hand \cite{moretti2011network, smeets2021modeling, sugishita2023social},
but manual annotations are time intensive and do not scale to large datasets.
Alternative approaches look for character co-occurrences in windowed units like  sentences or chapters \cite{way2018framework, evalyn2018analyzing, fischer2021social}.
Identifying co-occurrences is computationally lightweight, but their dependency on surface-level features limits their accuracy and applicability.
Neural networks have also been used more widely in recent years for this task \cite{nijila2018extraction, kim2019frowning, chen-etal-2020-mpdd, mellace2020temporal}. 
Specifically, \citet{piper-etal-2024-social} and \citet{zhao2024large} both use generative models to extract literary social networks, but their approaches are semi-supervised and thus not easily scaled, limiting their studies to datasets in the low hundreds of volumes.

\paragraph{Literary social networks in use.}

Literary social networks are often used to study particular character or character-relationship traits such as prominence \cite{masias2017exploring, sudhahar2013automated}, cooperativeness \cite{Chaturvedi_Srivastava_Daume_III_Dyer_2016}, relationship trajectory \cite{chaturvedi2017unsupervised, mellace2020temporal}, and relationship valence \cite{nijila2018extraction, kim2019frowning, piper-etal-2024-social}.
Some studies also use social networks to ground characters in particular locations \cite{lee2017shakespeare, lee2012extracting}.
Social networks are likewise useful for studying aspects of plot, including conflict \cite{smeets2021modeling}, narrative trajectory \cite{min2016network, moretti2011network}, textual genre \cite{agarwal2021genre, evalyn2018analyzing}, and text veracity \cite{sugishita2023social, volker2020imagined}.
They also provide data for studies comparing differences within a corpus \cite{fischer2021social}, over time \cite{algee2017distributed}, and between different social theories \cite{elson2010extracting, falk2016making, bonato2016mining, stiller2005weak, stiller2003small}.
However, these studies make use of relatively small corpora, limiting the statistical significance of their results.

\section{Methods}

\paragraph{Data.}

We draw volumes from the Project Gutenberg (PG) corpus \citep{hartProjectGutenberg1971}.
PG is an online collection of public domain literary volumes developed by volunteers.
It currently contains over 75,000 works, and continues to grow.
The size and historical breadth of the corpus makes it popular with researchers working in literary analysis \citep{brookeGutenTagNLPdrivenTool2015, reaganEmotionalArcsStories2016, piperBiodiversityNotDeclining2022} and corpus linguistics \citep{gerlachStandardizedProjectGutenberg2020}.

To create our dataset, we first download the full corpus from PG, resulting in 72,875 volumes totaling 25GB of raw text.\footnote{Our copy was obtained on September 29, 2024.}
We then supplement the limited metadata provided by PG (author and title) by aligning texts with 
MultiHATHI, an extended multilingual edition of the HathiTrust Digital Library catalog \citep{hamiltonMultiHATHICompleteCollection2023}, containing metadata such as publication date, language, and fiction/nonfiction status.
We use title and author edit distance \citep{levenshteinBinaryCodesCapable1966} to find the closest match for each PG text in MultiHATHI, only considering matches where the MultiHATHI title and author matches both exceed 80\% similarity.
This process yields 33,919 well-documented texts.

\paragraph{Model selection.}

We consider two qualities when selecting a suitable large language model (LLM) for generating social networks from arbitrary-length texts.
The first is maximum context length.
The longest work in our full corpus is 13,551,565 tokens (4,233,776 words) when tokenized with the SentencePiece-based Gemma 2 tokenizer \citep{kudoSentencePieceSimpleLanguage2018, teamGemma2Improving2024}.
For comparison, the mean word count of all volumes in our full corpus is 63,656 words ($P_{95}=170{,}601$).
Prompts of this magnitude can quickly exhaust the capacity of recent ``open weight'' LLMs, which most commonly offer context windows equal to or less than 128,000 tokens, despite the growing popularity of positional embedding modifications like RoPE and YaRN \citep{pengYaRNEfficientContext2023, suRoFormerEnhancedTransformer2023, jiangMistral7B2023, dubeyLlama3Herd2024}.

Our second consideration is support for structured output.
When we generate output for $\sim$70,000 documents from a stochastic language model, there is no default guarantee that the output will be consistent.
Recent methods for guaranteeing consistent output include grammar-constrained decoding, where tokens are selectively masked at sampling time according to some context-free grammar \citep{gerganovLlamacpp2024, microsoftGuidance2024, rickardReLLM2024, beurer-kellnerGuidingLLMsRight2024}.
A competing method is structured output, where the model emits JSON according to a JSON Schema passed at inference time \citep{shortenStructuredRAGJSONResponse2024}.
Along with larger context windows, proprietary models have made structured output a common feature.
From the pool of presently-available proprietary LLMs satisfying both these conditions, we select Google's \texttt{Gemini 1.5 Flash}, which features a context window of $1\times10^{6}$ tokens and supports structured output via JSON Schema \citep{teamGeminiFamilyHighly2024}.

\paragraph{Pipeline.}

To create an appropriate prompt for extracting social networks from texts, we turn to a public dataset released by \citet{massey2015annotating}. 
This dataset contains 2,170 annotated character relationships produced from 109 fictional narratives. 
Each character pair is labeled with three attributes: the valence of the relationship (positive, negative, or neutral) and two descriptors (``coarse-grained'' and ``fine-grained,'' with 3 and 30 possible labels respectively) further clarifying the relationship in terms of social function and connection (e.g., whether the characters are lovers).
\citet{massey2015annotating} release the annotation prompt they used for Mechanical Turk workers alongside the dataset.
We adapt their prompt for \texttt{Gemini 1.5 Flash} in JSON Schema, effectively requiring the model to return a JSON array of characters and their relationships for each text.\footnote{We provide an example prompt together with a list of rejected volumes \href{https://github.com/srhm-ca/pgn}{here.}}

\section{Results}

We use this pipeline to process the entire Project Gutenberg Corpus. 
It returns 71,836 networks from a total 72,875 volumes after omitting 1,039 volumes that fail to pass the Gemini API safe content filter.\footnote{Two works that fail to pass the safe content filter are Abraham Lincoln's \emph{Gettysburg Address} and an illustrated copy of Edgar Allen Poe's poetry. We do not investigate further the reasons for rejection in our corpus.}
Removing duplicate networks and malformed relationships (correcting attribute labels where possible) reduces this to 70,509 total networks, of which 29,346 (22,015 nonfiction and 7,331 fiction) have HathiTrust metadata available.

\paragraph{Validation.}

We assess the validity of our approach by comparing our Gemini-based pipeline against the human-annotated results reported in \citet{massey2015annotating}.
For each network in \citet{massey2015annotating}, we retrieve the original text and pass it through Gemini together with our prompt.
We additionally instruct the model to only return annotations for the character pairs pre-identified by \citet{massey2015annotating}.
We then calculate the ratio of true positive annotations over all annotations on a per-attribute basis to assess accuracy.

Identifying character networks and attributes is hard: \citet{massey2015annotating} report inter-annotator agreement rates of $\kappa=0.812, 0.744$ and $0.364$ for these tasks.
Our pipeline achieves a promising 81\% accuracy for valence and 74\% for ``fine category.''
However, Gemini does noticeably worse for ``coarse category'' annotations (55\%) despite the fact that each fine label is unique to a single coarse label (e.g., the fine label ``husband/wife'' implies the coarse label ``familial'').
We therefore make use of the coarse label annotations corresponding to the Gemini provided fine category labels in place of the originally produced values.

\section{Network Properties}

\begin{figure}[t]
    \centering
    \includegraphics[width=1\linewidth]{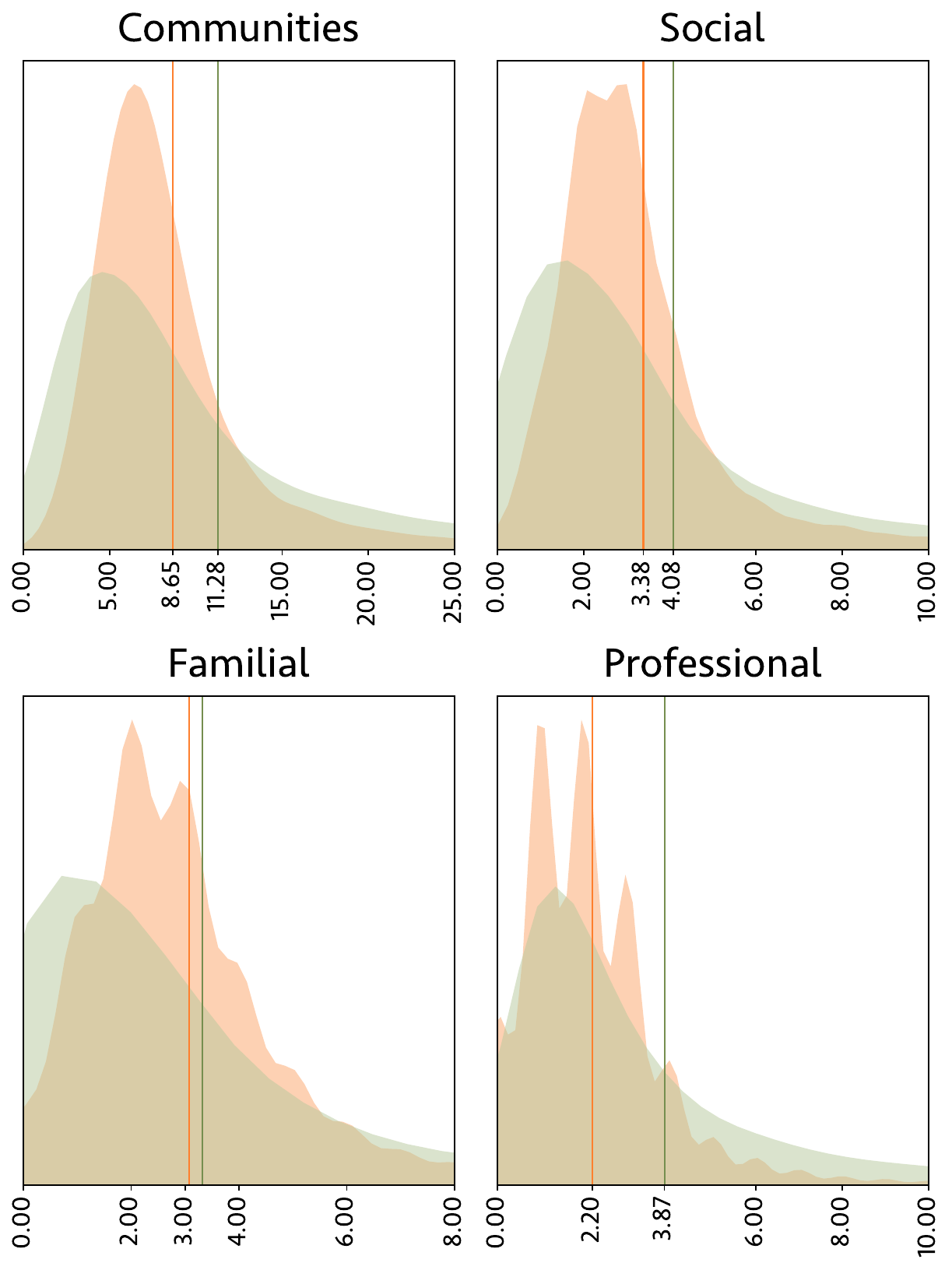}
    \caption{Density of community distributions for non-fiction (in green) and fiction texts (in orange). Vertical lines represent distribution means and graphs are truncated to show $90\%$ of the data. In all cases, fiction texts feature smaller mean community counts than non-fiction texts.}
    \label{fig:community_dists}
\end{figure}

Previous research has suggested that fictional worlds are smaller than nonfictional worlds.
For example, \citet{wilkens2024small} showed that fictional protagonists travel smaller distances, follow more routine paths, and more frequently spend time in domestic or private spaces than their nonfictional counterparts. 
With our new access to large-scale social network information, we can test whether similar distinctions hold between community structures.
In the same way that fictional characters travel less than do people in non-fiction, they may also participate in more tightly-knit social networks.

\paragraph{Network characteristics.} 
We test the validity of this hypothesis by assessing the network characteristics of the nonfiction and fiction volumes with metadata available in our dataset.
We find that nonfiction networks are on average significantly larger than fiction networks by both number of nodes (22.14 v.\ 42.69) and number of edges (27.42 v.\ 42.91).\footnote{All reported distinctions are significant under Welch's t-test at $p\!<\!10^{-9}$.}
Two other metrics of network complexity include the number of disconnected components (groups of characters that do not interact) and transitivity (the probability that two nodes that share a mutual connection are themselves connected).
Fictional networks have significantly fewer disconnected components (2.20 vs.\ 5.12) and their mean transitivity is significantly larger (0.22 vs.\ 0.12) than nonfiction networks. Thus, we see that fiction networks are smaller and more clustered than nonfiction networks.

\paragraph{Community detection.}

While completely disconnected sub-graphs are easy to identify, there are also often denser \textit{communities} embedded in larger graphs.
Since our networks can contain multiple edges between two nodes representing different relationships, we first divide the full network into three networks (familial, social, professional).
We then use the Louvain method \cite{blondel2008fast} to partition each graph into communities, with no pre-set number of communities per graph.

The mean number of communities in non-fiction networks is indeed larger than in fiction networks (11.28 vs.\ 8.65).\footnote{This method of calculating community count assumes that a majority of communities will only contain relationships of a single edge type. Results with communities drawn from the entire graph are similar.}
This is true even if we look only at social (3.38 vs.\ 4.08), familial (3.07 vs.\ 3.32), or professional (2.20 vs.\ 3.87) communities. 
Figure \ref{fig:community_dists} shows the distribution of community counts for fiction and non-fiction.
Fictional networks tend to have a more consistent number of communities, while non-fiction networks have a wider range; if a work has very few or very many communities, it is more likely to be non-fiction.

\paragraph{Relationship types.}
We observe that fiction networks consist of a significantly larger proportion of social (48.08\% v.\ 39.44\%) and familial (30.07\% v.\ 21.15\%) relationships on average, whereas nonfiction networks have a larger average proportion of professional relationships (39.41\% v.\ 21.86\%). This aligns with \citet{wilkens2024small}'s finding that fictional characters spend more time in domestic and private spaces.

\section{Conclusion}

This work presents a novel dataset containing 70,509 high-quality social networks extracted from fiction and nonfiction narratives.
It additionally includes metadata for 29,346 texts written between 1800 and 1999 in 58 languages.
We release this resource to support researchers in the humanities and social sciences studying the development of social worlds over time, and the work of behavioral scientists who seek to understand how cognitive models of social communities compare with real-world social communities.

Our dataset-construction process also contributes to a growing literature on adapting annotation task instructions for language models.
Our results demonstrate that we can use LLMs to generate large-scale datasets for complicated and nuanced annotations on volume-length data.
We find that constrained output such as JSON Schema is critical to maintaining consistency and compatibility at scale.
We also observe that more concrete, descriptive annotations are more successful than more abstract annotations, even when these appear logically identical to humans.

\paragraph{Next steps.}

While our dataset is a step forward for researchers studying social networks, there remains room for progress.
Generative language models improve over older social network extraction methods based on surface-level features, but we need better open and locally runnable alternatives to inefficient and costly proprietary models.
These models have low interpretability, do not permit unlimited token lengths, and block content considered inappropriate for opaque reasons that may be inappropriate for historical data. 
We similarly lack good evaluation methods.
Modeling social networks as graphs is an inherently \emph{interpretive} act, sometimes literally so in the context of literary data.
To that end, social networks may change depending on the perspective of the narrator.
Our current method does not allow these perspectives be reflected in graph structures.
We believe future research should consider further methods for assessing the validity of extracted social networks.

\section*{Limitations}
We note three primary limitations impacting this work.
First, our textual data is sourced from predominantly European authors.
Because it is an American project, the vast majority of volumes in Project Gutenberg are written in English.
While the dataset does contain volumes written in at least 58 languages, the three most dominant are English, French, and German.
The second limitation is that \texttt{Gemini 1.5 Flash} has a maximum context window of one million tokens.
This means our pipeline could not process $\sim$374 volumes whose token counts exceed this maximum (although we note that the average text in our dataset contains 63,656 words, two orders of magnitude below this maximum).
Finally, \texttt{Gemini 1.5 Flash} can only emit a maximum of 8,000 tokens in one API call.
Our results indicate that some volumes contain social networks exceeding this maximum, suggesting some networks included in our dataset are incomplete.

\section*{Acknowledgments}
We thank Surendra Ghentiyala, Jon Kleinberg, and Andrew Piper for their comments throughout this project. This work was supported by NEH grant HAA-290374-23, AI for Humanists, granted to Matthew Wilkens and David Mimno.

\bibliography{anthology,custom}
\bibliographystyle{acl_natbib}

\end{document}